\documentclass[runningheads]{llncs}
\usepackage[T1]{fontenc}
\usepackage{graphicx,verbatim}
\usepackage{amssymb}
\usepackage{amsmath}
\usepackage{booktabs}
\usepackage{subcaption}
\usepackage{multirow}
\usepackage{amsfonts}
\usepackage[bookmarks]{hyperref}
\usepackage{booktabs} 
\usepackage{multirow}
\usepackage{subfiles} % Best loaded last in the preamble

\begin{document}
\title{HyDA:  Hypernetworks for Test Time Domain Adaptation in Medical Imaging Analysis}
\titlerunning{Hypernetworks for Test Time Domain Adaptation in Medical Imaging}
%

% \orcidID{0009-0000-3941-5017} \orcidID{0000-0003-2532-5875}}
\author{Doron Serebro\inst{1} \and 
Tammy Riklin-Raviv\inst{1}}
\authorrunning{D. Serebro et al.}

\institute{Ben-Gurion University of The Negev \and
\email{doronser@post.bgu.ac.il} \and 
preprint version}

\maketitle              % typeset the header of the contribution

\begin{abstract}
Medical imaging datasets often vary due to differences in acquisition protocols, patient demographics, and imaging devices. These variations in data distribution, known as domain shift, present a significant challenge in adapting imaging analysis models for practical healthcare applications.

Most current domain adaptation (DA) approaches aim either to align the distributions between the source and target domains or to learn an invariant feature space that generalizes well across all domains. However, both strategies require access to a sufficient number of examples, though not necessarily annotated, from the test domain during training. This limitation hinders the widespread deployment of models in clinical settings, where target domain data may only be accessible in real time.

In this work, we introduce HyDA, a novel hypernetwork framework that leverages domain characteristics rather than suppressing them, enabling dynamic adaptation at inference time. Specifically, HyDA learns implicit domain representations and uses them to adjust model parameters on-the-fly, effectively interpolating to unseen domains. We validate HyDA on two clinically relevant applications—MRI brain age prediction and chest X-ray pathology classification—demonstrating its ability to generalize across tasks and modalities.
Our code is available at TBD.
% Authors must provide keywords and are not allowed to remove this Keyword section.
\end{abstract}
\section{Introduction}
Deep learning significantly advanced medical image analysis by enabling accurate detection, classification, segmentation, and predictive modeling. However, for practical healthcare applications, models must adapt to variations in imaging protocols, scanner types, and patient demographics, which create differing data distributions between training and test sets. This discrepancy, known as domain shift, remains a significant barrier to robust model performance. 

Current domain adaptation (DA) techniques generally aim to either align the distributions between source and target domains or learn a consistent feature space across different domains. Yet, both approaches depend on having enough target domain samples during training, whether annotated or not. This dependency poses a challenge for the deployment of models in clinical settings, where the target data may only be available at the time of the test. 
% \cite{uda-gu2024} %\cite{uda-med-survey}
In this work, we propose HyDA, a novel hypernetwork framework that exploits domain characteristics rather than discarding them, enabling dynamic adaptation during both training and inference. Specifically, HyDA learns implicit domain representations that are used to generate weights and biases for a primary network on-the-fly, effectively interpolating to unseen domains.
HyDa is task and modality agnostic, making it easily integrable into various medical imaging applications. We showcase its generality and robustness for two clinically relevant tasks—chest X-ray pathology classification and MRI brain age prediction, demonstrating superior performance over baseline and other domain adaptation techniques.  

\section{Related Works}
\textbf{Domain Adaptation Methods.} Unsupervised domain adaptation (UDA) addresses shifting data distributions between source and target domains. One prominent approach is domain adversarial learning, as exemplified by Domain-Adversarial Neural Networks (DANN)\cite{Ganin-uda-dann-2016}, while MDAN (Multi-Domain Adversarial Network) extends this idea to multiple source domains\cite{Zhao-uda-mdan-2018}. Another line of works focuses on invariant feature learning; for example, Deep CORrelation ALignment (CORAL) minimizes domain discrepancy by aligning the second-order statistics of source and target features~\cite{sun2016deep}. \\
Recently, transformer-based methods have gained traction for their self-attention capabilities. TransDA leverages domain-specific tokens and cross-attention to align features unsupervisedly~\cite{yang2021transformer}. Similarly, AdaptFormer~\cite{chen2022adaptformer} and DAFormer~\cite{hoyer2022daformer} integrate lightweight adapter modules within a Vision Transformer framework to modulate representations based on domain cues while maintaining a shared global representation.

Test-time domain adaptation (TTDA) techniques have a key advantage over the methods mentioned above: they do not require target data during training and can adapt models on-the-fly during inference. For example, TENT (Test Time Entropy Minimization) adjusts model parameters via entropy minimization, which works well for multi-class classification tasks with clear output probabilities~\cite{tta-tent}. MEMO stabilizes adaptation under distribution shifts using augmentations \cite{tta-memo}. Although not strictly a TTDA method, SHOT (Source Hypothesis Transfer) adapts to target data without requiring source samples~\cite{tta-shot} by relying on pseudo-labeling and entropy minimization. However, the reliance on entropy may limit their applicability to tasks such as regression or multi-label classification without modifications.

\textbf{Hypernetworks.}
First introduced by Ha et al.\cite{hypernetworks}, hypernetworks are neural networks that generate weights and biases for primary networks, dynamically creating a unique set of parameters for each input. Their effectiveness has been demonstrated in various tasks, including 3D shape reconstruction\cite{hyper-3d-rec}, federated learning~\cite{hyper-fed}, and medical image segmentation~\cite{hyper-med-seg}. Aharon et al.\cite{hyper-denoise} showed that hypernetworks can interpolate by conditioning an image denoising model on expected noise variance, while Duenias et al.\cite{hyperfusion} used them to condition medical image analysis on tabular data. Building on these ideas, we show that hypernetworks can be applied to medical imaging domain adaptation by generating weights from domain features, effectively interpolating across the domain space.
\section{Method}
\begin{figure}[t!]
    \centering
    \includegraphics[scale=0.6]{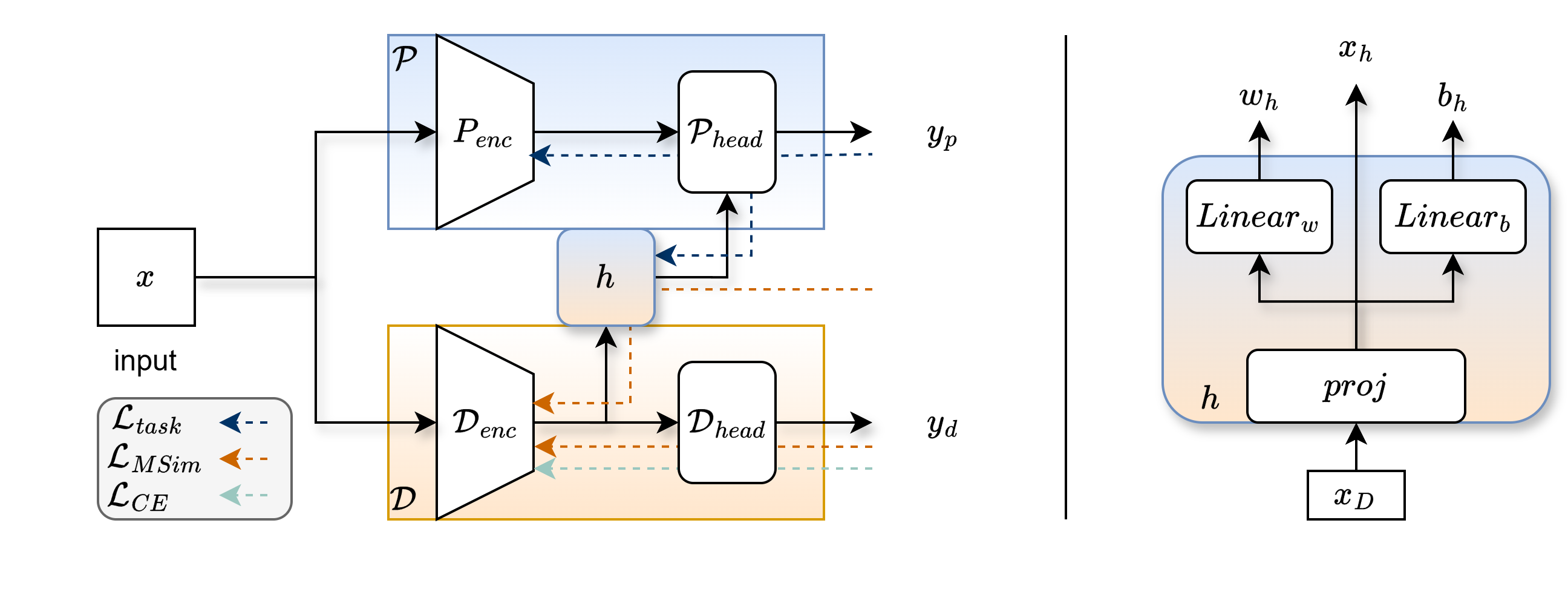}
    \caption[]{
    The proposed HyDA framework (left) is composed of a hypernetwork $h$ (right), a primary network $\mathcal{P},$ and a domain classifier $\mathcal{D}.$ The hypernetwork generates weights and biases to the primary's head $\mathcal{P}_{head}$ based of the domain feature vector $X_D$ provided by the domain encoder $\mathcal{D}_{enc}$. Other weights in the system are internal and are learned via back-propagation using a regularized task dependent $\mathcal{L}_{RT}$, classification $\mathcal{L}_{CE}$ or/and multi-similarity $\mathcal{L}_{MSim}$ loss functions as illustrated by the dashed arrows. 
    }
    \label{fig:method-overview}
\end{figure}

Our proposed HyDA framework, illustrated in Fig.~\ref{fig:method-overview}, is composed of a primary network $\mathcal{P}$ that could have any architecture addressing any medical imaging analysis task; a hypernetwork $\mathit{h}$ and a domain classifier $\mathcal{D}.$ Being trained on datasets from different source domains - the classifier learns implicit domain features that are mapped by $\mathit{h}$ to sets of weights and biases. These parameters, termed \emph{external parameters} are transferred to a subset of layers in $\mathcal{P}.$ 

Formally, let $x \in \mathbb{R}^{d}$, denote a $d-$dimentional input image ($d\in\{2,3\}$). 
We define by $\mathcal{D}_{enc}$ and $\mathcal{D}_{head}$ the domain encoder and domain head, respectively, which together compose the classifier $\mathcal{D}$. 
%In a similar manner, the primary network $\mathcal{P}$ can be decomposed into task encoder %$\mathcal{P}_{enc}$ and task head $\mathcal{P}_{head}$. 

The classifier is trained to predict the domain of a training image $x$ as follows:
\begin{equation}
    y_D = \mathcal{D}_{head}(\mathcal{D}_{enc}(x))
\end{equation}
where $y_D$ is the label of a source domain. We assume the availability of at least two source domains. If deployed separately, the trained domain encoder maps any input $x$ into a domain feature vector, i.e.,
\begin{equation}
    x_D = \mathcal{D}_{enc}(x)
\end{equation}
where $x_D \in f_D$ is the domain feature vector of $x$ and $f_D$ denotes the domain feature space.
The primary network $\mathcal{P}$ which is trained to predict an output $y_P$ for $x$ can be formalized as follows: 
\begin{equation}
    y_P = \mathcal{P}_{head}(\mathcal{P}_{enc}(x), \mathit{h}(D_{enc}(x))), 
\end{equation}
where, $\mathcal{P}_{enc}$ denotes the \emph{internal} layers in $\mathcal{P}$ which are trained through a standard back propagation process and $\mathit{h}(D_{enc}(x))$ defines its \emph{external} domain-aware weights and biases - generated by the hypernetwork $\mathit{h}.$ 

We hypothesize that during inference, feature vectors of a target domain $x_D$, unseen by the domain encoder before, are well embedded within the domain feature space, $f_D,$ and can be represented as linear combinations of training domain features. We aim to optimize the hypernetwork such that the metric capturing inter- and intra-domain relationships within $f_D$ is preserved in the external primary network parameters. Once optimally converged, HyDA can interpolate to new target domains at test time.

\subsection{Domain Conditioning Hypernetwork}
The hypernetwork maps the domain embedding $x_D$ to weights and biases $(w_h, b_h)$
for the external primary network layers. 
Let $N,O,B$ denote the layer's input, output and batch size, respectively.  
In a standard linear layer, the output $\psi \in \mathbb{R}^{(B,O)}$ is computed as:  
\[
\psi = \chi * w, \quad w \in \mathbb{R}^{(N,O)}
\]
where \( \chi \in \mathbb{R}^{(B,N)} \) is the input batch and $*$ is matrix multiplication.  
A hyper-linear layer instead assigns a unique weight matrix to each batch element:  
\[
\psi_i = \chi_i * w_h^i, \quad w_h^i \in \mathbb{R}^{(N,O)} \quad i =1, \ldots B 
\]

The hypernetwork is flexible and can be implemented in various ways. For simplicity, we use a single linear layer followed by a ReLU activation, which is sufficient to generate effective domain-aware weights for the primary network.

To ensure stable convergence, we initialize the hypernetwork weights as in Chang et al.~\cite{hyper-weights_init} such that the input variance is preserved in the primary network. We also regularize the weights using the $l_2$ norm. 
\subsection{Loss Functions}
The hypernetwork and the internal primary network layers are trained in an end-to-end manner with a regularized loss function as follows:
\begin{equation}
   \mathcal{L}_{RT} = \mathcal{L}_{task} + \lambda_{BP}\left\| w_{BP}\right\|_2  + \lambda_h\left\|w_h\right\|_2
\end{equation}
where, RT stands for regularized task, BP for backpropagation, $\mathcal{L}_{task}$ is a task-dependent loss (e.g. cross-entropy for classification, MSE for regression), $w_{BP}$ denote the union of the hypernetwork's and the internal primary network's  weights, $w_h$ are the exterrnal primary network weights, generated by the hypernetwork, and $\lambda_{BP}, \lambda_h$ are their corresponding coefficients.
The domain classifier is trained using the following loss:
\begin{equation}
    \mathcal{L}_D = \mathcal{L}_{CE} +\alpha \mathcal{L}_{MSim} + \lambda_D\left\|w_D\right\|_2 
\end{equation}
where $\mathcal{L}_{CE}$ is the cross-entropy loss, $\mathcal{L}_{MSim}$ is multi-similarity loss as in Wang et. al.~\cite{msim_loss}, $w_D$ are the domain network's weights and $\alpha, \lambda_D$ are coefficients. The supervised CE loss $\mathcal{L}_{CE}$ aims to correctly classify the input into source domains, while the contrastive, multi-similarity loss encourages the separation of embedded domain feature vectors into different domain-aware clusters.    

The multi-similarity loss ($\mathcal{L}_{MSim}$ ) also supports the hypernetwork training - allowing it to maintain domain-specific representation of the weights and biases it generates for the primary network.

\section{Experiments and Results}
We demonstrate the proposed HyDA framework on two medical imaging analysis tasks - chest X-ray pathology classification, and MRI brain age prediction.
\subsection{Chest X-ray Pathology Classification}
We trained our model for multi-label classification on chest X-ray scans from three publicly available datasets, comparing HyDA to a baseline with no adaptation, a UDA method (soft MDAN~\cite{Zhao-uda-mdan-2018}), and a TTDA method (TENT~\cite{tta-tent}). In both cases, the domain classifier was pre-trained for robust initialization.\\
\textbf{The Data.} We use the NIH~\cite{cxr-nih}, CheXpert~\cite{cxr-chexpert}, and VinDr~\cite{cxr-vindr} datasets, selecting five classes—Atelectasis, Cardiomegaly, Consolidation, Effusion, and Pneumothorax—that are common across all three, resulting in a combined dataset of 90,570 X-ray scans. 

\begin{figure}[ht]
    \centering
    \begin{minipage}{0.24\textwidth}
        \centering
        \includegraphics[width=\linewidth]{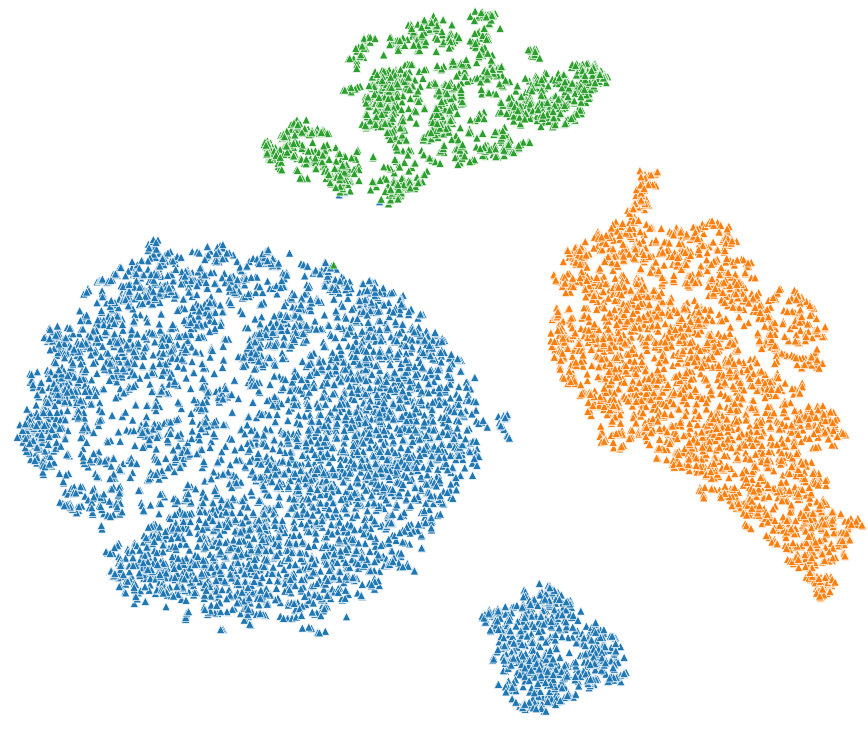}
        \subcaption{All domains}
    \end{minipage}
    \begin{minipage}{0.24\textwidth}
        \centering
        \includegraphics[width=\linewidth]{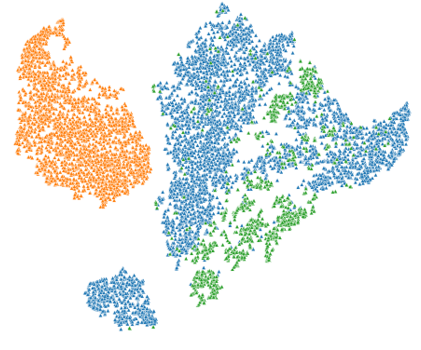}
        \subcaption{w/o NIH}
    \end{minipage}
    \begin{minipage}{0.24\textwidth}
        \centering
        \includegraphics[width=\linewidth]{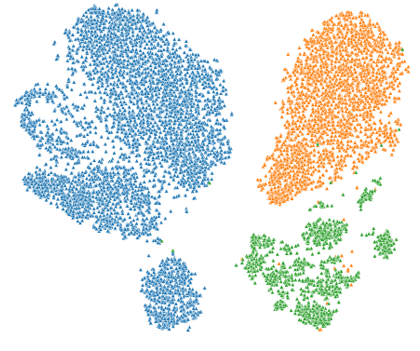}
        \subcaption{w/o CheXpert}
    \end{minipage}
    \begin{minipage}{0.24\textwidth}
        \centering
        \includegraphics[width=\linewidth]{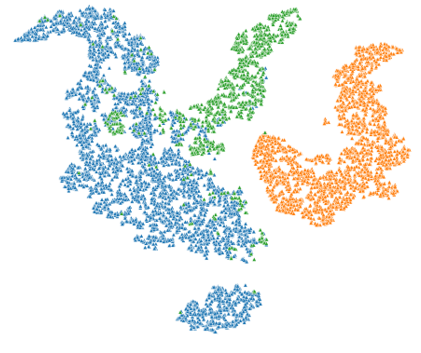}
        \subcaption{w/o VinDr}
    \end{minipage}
    \caption{t-SNE projections of domain feature in fully supervised and leave-one-out settings. The plots illustrate the embedding of previously unseen domains in the learned domain feature space : (a) All domains training with respect to training  (b) w/o NIH (blue) (c) w/o CheXpert (orange) and (d) w/o VinDr (green).}
    \label{fig:cxr-dom-clf-tsne}
\end{figure}

\noindent{\textbf{Implementation Details.}} We fine-tuned a DenseNet121 model pre-trained on ImageNet, replacing its input and output layers to process single-channel images and output five classes, following prior work~\cite{cxr-chexnet,cxr-xrv}. The domain classifier is a simple CNN with four convolution blocks and a linear classification layer, while the hypernetwork is a multi layer perceptron (MLP) that generates a set of weights and biases for the primary network (DenseNet). Both baseline and HyDA models were trained for 150 epochs using the AdamW optimizer (learning rate: 1e-3, weight decay: 0.05) with a cosine annealing scheduler (minimum learning rate: 1e-6). \\
\noindent\textbf{Results.}
Table~\ref{tab:cxr-results} reports chest X-ray experiment results in terms of area under curve (AUC). HyDA outperforms the baseline in both fully supervised and leave-one-out settings. Notably, the improvement correlates with the separability of domain features that were not seen in training (see Fig.~\ref{fig:cxr-dom-clf-tsne}); domains with well-clustered features (CheXpert and VinDr) show larger gains compared to NIH.
\begin{table}[tb!]
    \centering
    \small
    \renewcommand{\arraystretch}{1.1}
    \setlength{\tabcolsep}{4pt}
    \begin{tabular}{lp{1.2cm}cccccc}
        \hline
        \multirow{2}{*}{ \shortstack{\textbf{Target} \\ \textbf{Domain}}} & \multirow{2}{*}{\textbf{Method}} & \multicolumn{5}{c}{\textbf{Pathologies (AUC) ↑}} & \multirow{2}{*}{\textbf{Avg.}} \\
        \cline{3-7}
        & & Atel. & Cardio. & Cons. & Eff. & Pneu. &  \\
        \hline
        \multirow{4}{*}{\shortstack[l]{\textbf{-}}} 
        & Baseline  & 0.85  & 0.95  & 0.86  & 0.94  & 0.87  & 0.89 \\
        & MDAN      & 0.86  & 0.96  & 0.86  & 0.94  & 0.88  & 0.90 \\
        & HyDA      & \textbf{0.87}  & \textbf{0.97}  & \textbf{0.86}  & \textbf{0.94}  & \textbf{0.89}  & \textbf{0.91} \\
        \hline
        \multirow{4}{*}{\shortstack[l]{\textbf{NIH}}} 
        & Baseline & \textbf{0.70}  & 0.81  & \textbf{0.76}  & 0.86  & 0.77  & 0.78 \\
        & MDAN     & 0.67  & 0.89  & 0.76  & 0.86  & 0.77  & 0.79 \\
        & TENT     & 0.61  & 0.70  & 0.64  & 0.81  & 0.67  & 0.69 \\
        & HyDA     & 0.68  & \textbf{0.89}  & 0.75  & \textbf{0.88}  & \textbf{0.79}  & \textbf{0.80} \\
        \hline
        \multirow{4}{*}{\shortstack[l]{\textbf{CheXpert}}} 
        & Baseline & 0.81  & \textbf{0.86}  & 0.73  & 0.87  & 0.74  & 0.80 \\
        & MDAN     & 0.77  & 0.76  & 0.71  & 0.84  & 0.72  & 0.76 \\
        & TENT     & 0.76  & 0.86  & 0.77  & 0.89  & 0.76  & 0.81 \\
        & HyDA     & \textbf{0.82}  & 0.85  & \textbf{0.82}  & \textbf{0.89}  & \textbf{0.74}  & \textbf{0.82} \\
        \hline
        \multirow{4}{*}{\shortstack[l]{\textbf{VinDr}}}  
        & Baseline & 0.60  & 0.76  & 0.85  & 0.88  & 0.91  & 0.80 \\
        & MDAN     & \textbf{0.68}  & 0.82  & 0.88  & 0.87  & 0.89  & 0.83 \\
        & TENT     & 0.51  & 0.72  & 0.80  & 0.74  & 0.86  & 0.73 \\
        & HyDA     & 0.66  & \textbf{0.87}  & \textbf{0.93}  & \textbf{0.89}  & \textbf{0.92}  & \textbf{0.85} \\
        \hline
    \end{tabular}
    \caption{Chest X-ray classification results measured by AUC. Pathologies abbreviations: Atel (Atelectasis), Cardio (Cardiomegaly), Cons (Consolidation), Eff (Effusion), Pneu (Pneumothorax). Each group compares different models on the same target domain. Best results in bold.}
    \label{tab:cxr-results}
\end{table}

\noindent\textbf{Ablation Study.}
Table~\ref{tab:cxr-loss-ablation} presents an ablation study of the proposed loss functions. The results highlight the contribution of each loss component to achieving the best possible performance.

\subsection{Brain Age Prediction}
To further assess our method demonstrating its being task agnostic we evaluated its performances for age prediction from brain MRI scans.   
\begin{table}[tb!]
    \centering
    \begin{tabular}{cccccc}
        \toprule
        $\mathcal{L}_{CE}$ & $\mathcal{L}_{MSim}^D$  & $\mathcal{L}_{MSim}^h$ & \textbf{NIH} & \textbf{CheXpert} & \textbf{VinDr} \\
        \midrule
        \checkmark & & & 0.72 (0.11) & 0.79 (0.05) & 0.81 (0.13) \\
        \checkmark & \checkmark & & 0.76 (0.09) & 0.81 (0.06) & 0.83 (0.10) \\
        \checkmark & \checkmark & \checkmark & \textbf{0.80 (0.08)} & \textbf{0.82 (0.05)} & \textbf{0.85 (0.10)} \\
        \bottomrule
    \end{tabular}
    \caption{Ablation study of the loss terms. Each row represents an incremental combination of loss terms, including domain classifier's cross-entropy (CE) $\mathcal{L}_{CE}$ and multi-similarity (MSim) $\mathcal{L}_{MSim}^D$ loss functions as well as hypernetworks' MSim loss $\mathcal{L}_{MSim}^h$. Average AUC results (std in brackets) of the target domain for each of the three datasets are reported.}
    \label{tab:cxr-loss-ablation}
\end{table}
%\vspace{-.5cm}
\\
\noindent\textbf{The Data.} We used 19 brain MRI datasets containing 26,691 scans. The scans were preprocessed using the workflow in Levakov et. at. \cite{brain_age-levakov}.

\noindent\textbf{Implementation Details.}
Our primary network follows Levakov et. al. \cite{brain_age-levakov} 3D CNN comprised of 4 convolution blocks followed by a 4-layered-MLP. The domain classifier follows a similar architecture, with fewer parameters (refer to our code for further details).
The model was trained using AdamW optimizer with a learning rate of $1e-4$, weight decay of $0.05$ and a cosine annealing learning rate scheduler with a minimum learning rate of $1e-6$ for 150 epochs.

\begin{figure}[ht]
    \centering
    \begin{subfigure}{0.35\textwidth}
        \centering
        \includegraphics[width=\linewidth]{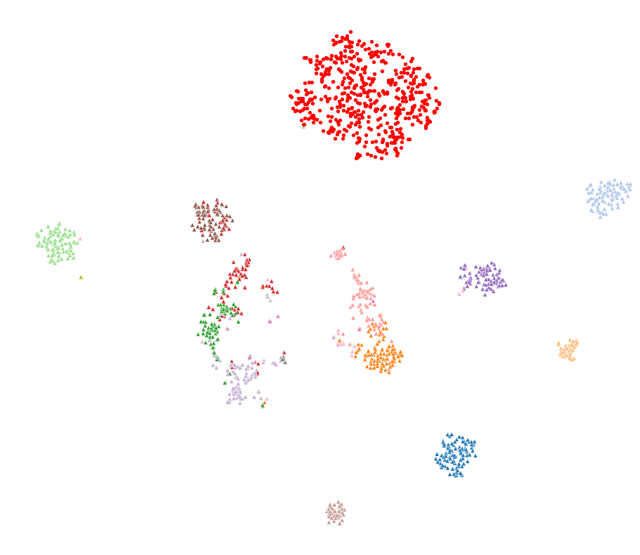}
        \caption{All domains training}
    \end{subfigure}
    \hfill
    \begin{subfigure}{0.35\textwidth}
        \centering
        \includegraphics[width=\linewidth]{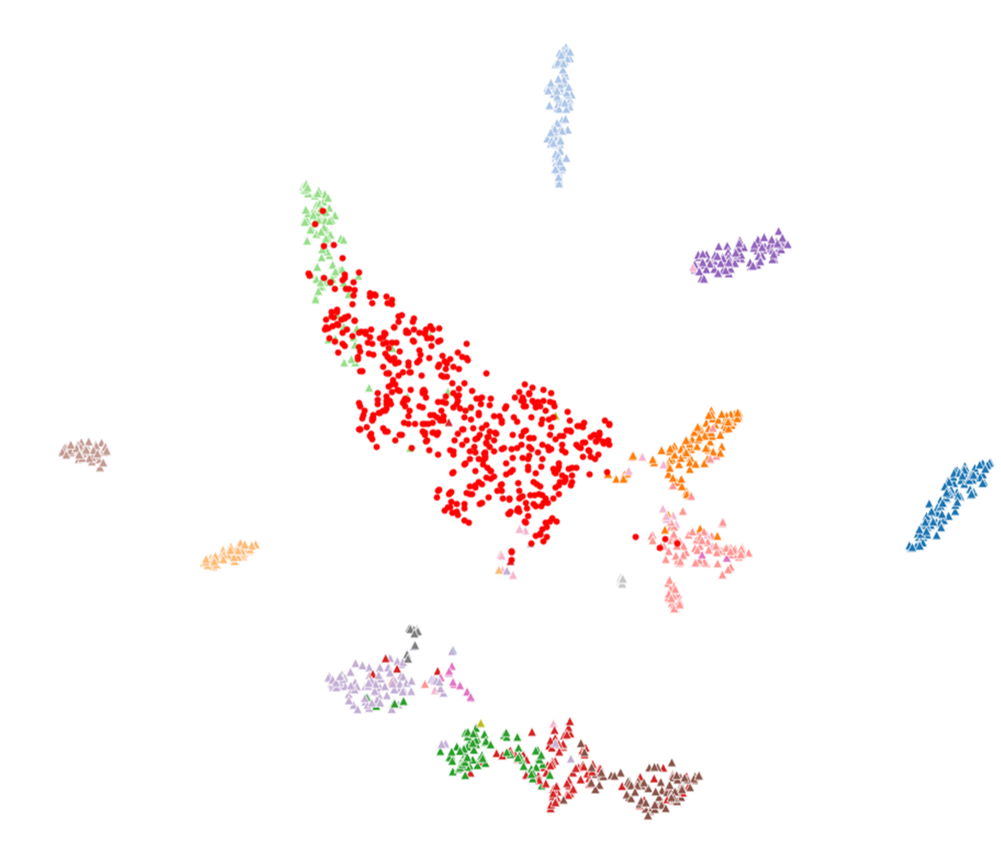}
        \caption{All domains w/o Camcan}
    \end{subfigure}
    \hfill
    \begin{subfigure}{0.13\textwidth}
        \centering
        \includegraphics[width=\linewidth]{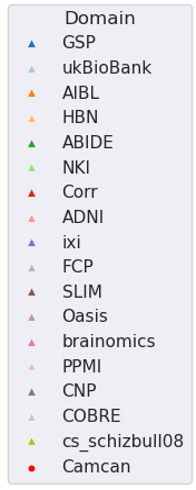}
    \end{subfigure}
    \caption{t-SNE projections of domain feature vectors $x_D$, Camcan examples are in red. The plots show embedding when samples of all domains (a) or all but Camcan (b) are available during training.}
    \label{fig:mri-camcan-tsne}
\end{figure}

\noindent\textbf{Results.}
Table~\ref{tab:brain_age_results} shows brain age prediction results. Notably, HyDA improves over the baseline for both supervised a leave-one-out settings. These results demonstrate HyDA's ability to learn meaningful domain representations, interpolate to unseen domains and utilize the domain features to adapt the model on-the-fly. The ability to interpolate is demonstrated in the t-SNE plot in Fig.~\ref{fig:mri-camcan-tsne} which shows how samples from a previously unseen domain are well embedded in between feature vectors from domains used for training. 

\begin{table}[tb!]
    \centering
    \renewcommand{\arraystretch}{1.2}
    \resizebox{\textwidth}{!}{
    \begin{tabular}{lccccccccccc}
        \toprule
        \cmidrule(lr){2-11}
        \textbf{Model} & \textbf{CNP}\cite{poldrack2016CNP} & \textbf{NKI} \cite{nooner2012nkiRS} & \textbf{ixi} \cite{heckemann2003ixi}& \textbf{Oasis} \cite{marcus2007OASIS} & \textbf{ABIDE} \cite{di2014ABIDE} & \textbf{ADNI} \cite{jack2008ADNI} & \textbf{AIBL} \cite{ellis2009AIBL} & \textbf{PPMI} \cite{MAREK2011PPMI} & \textbf{Camcan} \cite{shafto2014CamCAN} & \textbf{SLIM} \cite{liu2017SLIM} & \textbf{Avg. (std)} \\
        \midrule
        \multicolumn{12}{c}{\textbf{Fully Supervised (Validation MAE) ↓}} \\
        \midrule
        Baseline & 3.11 & 3.01 & 3.54 & \textbf{3.29} & 2.09 & \textbf{2.80} & \textbf{2.74} & 4.23 & 3.35 & 0.47 & 2.86 (0.96) \\
        HyDA & \textbf{2.39} & \textbf{2.92} & \textbf{3.22} & \textbf{3.29} & \textbf{1.74} & 3.04 & 2.94 & \textbf{3.94} & \textbf{3.21} & \textbf{0.37} & \textbf{2.71 (0.95)} \\
        \midrule
        \multicolumn{12}{c}{\textbf{Leave-One-Out (Test MAE) ↓}} \\
        \midrule
        Baseline & 3.36 & 3.90 & 4.41 & 5.40 & 3.25 & \textbf{4.31} & 3.56 & \textbf{4.15} & 3.50 & 1.44 & 3.73 (0.97) \\
        HyDA & \textbf{2.86} & \textbf{3.44} & \textbf{4.14} & \textbf{5.20} & \textbf{3.16} & 4.48 & \textbf{3.45} & 4.24 & \textbf{3.35} & \textbf{1.34} & \textbf{3.57 (1.00)} \\
        \bottomrule
    \end{tabular}
    }
    \caption{Brain age prediction results in fully supervised (validation MAE) and leave-one-out (test MAE) settings. Best scores are in bold.}
    \label{tab:brain_age_results}
\end{table}

\noindent\textbf{Ablation Study}
We evaluated the robustness of the HyDA by testing different configurations of the primary network's MLP head, which consists of four layers, three of which can be external, having their weights generated by the hypernetwork. Table~\ref{tab:mri-layers-ablation} shows that replacing some internal layers with domain-specific (external) ones improves performance over the baseline, regardless of which layers are adapted. The best performance is achieved by combining both; relying solely on task-specific weights limits generalization to unseen domains, while using only domain-specific weights loses critical task-related information.
% \begin{table}[tb!]
%     \centering
%     \renewcommand{\arraystretch}{1.2}
%     \begin{tabular}{ccccl}
%         \hline
%         \textbf{Layer 1} & \textbf{Layer 2} & \textbf{Layer 3} & \textbf{Average (std)} \\
%         \hline
%         & & & 4.57 (0.62) \\
%         \checkmark & & & 4.36 (0.55) \\
%         & \checkmark & & 4.43 (0.70) \\
%         & & \checkmark & 4.28 (0.72) \\
%         & \checkmark & \checkmark & 4.26 (0.72) \\
%         \checkmark & \checkmark & \checkmark & 4.68 (0.73) \\
%         \hline
%     \end{tabular}
%    \caption{Hypernetwork external layer configuration - ablation study. Each configuration was trained on three target domains (NKI, ixi, Oasis), and results are reported as mean (std) target domain MAE.
% \label{tab:mri-layers-ablation}}    
% \end{table}

\begin{table}[tb!]
    \centering
    \renewcommand{\arraystretch}{1.2}
    \begin{tabular}{ccccl}
        \hline
        \textbf{Layer 1} & \textbf{Layer 2} & \textbf{Layer 3} & \textbf{Average (std)} \\
        \hline
        & & & 4.16 (0.26) \\
        \checkmark & & & 3.99 (0.20) \\
        & \checkmark & & 3.97 (0.32) \\
        & & \checkmark & 3.79 (0.21) \\
        & \checkmark & \checkmark & 3.79 (0.35) \\
        \checkmark & \checkmark & \checkmark & 4.17 (0.11) \\
        \hline
    \end{tabular}
   \caption{Hypernetwork external layer configuration - ablation study. Each configuration was trained on two target domains (NKI, ixi), and results are reported as mean (std) target domain MAE.}
\label{tab:mri-layers-ablation}
\end{table}

\vspace{-.2cm}
\section{Conclusions}
We introduced HyDA, a hypernetwork-based framework that rethinks test-time domain adaptation in medical imaging by embracing domain variability rather than eliminating it. By learning implicit domain representations and dynamically generating model parameters at test time, HyDA effectively tailors predictions for each input based on its domain characteristics. 

Experimental evaluations on chest X-ray pathology classification and MRI brain age prediction demonstrate that HyDA outperforms traditional domain-invariant methods and existing test-time adaptation techniques. The framework’s ability to interpolate between domains, as revealed through t-SNE visualizations, confirms that leveraging domain-specific cues leads to more robust and generalizable models. Moreover, the task-agnostic design and compatibility with various architectures make HyDA a versatile solution for a wide range of clinical applications.
Overall, HyDA offers a promising pathway toward more reliable and adaptable medical imaging analysis, paving the way for models that can seamlessly adjust to real-world variations in data acquisition without the need for extensive target domain training.

%We introduced HyDA, a novel hypernetwork-based framework for %unsupervised domain adaptation in medical imaging. Unlike 5
%traditional domain adaptation methods that seek to eliminate domain-%specific variations, HyDA leverages domain information to %dynamically adjust model parameters at inference time. This enables %on-the-fly adaptation to unseen domains without requiring target %domain data during training.

%Through evaluations on MRI brain age prediction and chest X-ray %pathology classification, we demonstrated HyDA’s effectiveness in %improving generalization across tasks and modalities. Our results %highlight HyDA’s ability to interpolate between domains, leading to %improved robustness in clinical applications.
%As a task- and modality-agnostic framework, HyDA can be integrated %into various medical imaging pipelines, offering a scalable %solution for overcoming domain shift in real-world healthcare %settings.
\bibliographystyle{splncs04}
\bibliography{bibliography}
\end{document}